\documentclass[10pt,twocolumn,letterpaper]{article}
\usepackage[pagenumbers]{cvpr}
\usepackage{url}
\usepackage{graphicx}

\usepackage{booktabs}       
\usepackage{amsfonts}       
\usepackage{nicefrac}       
\usepackage{xcolor}         

\usepackage{amsmath}
\usepackage{amssymb}
\usepackage{graphicx}
\usepackage{multirow}
\usepackage{bm}
\usepackage{array}
\usepackage{makecell}
\usepackage{adjustbox}
\usepackage[table]{xcolor}
\usepackage{pifont}
\usepackage{algorithm}
\usepackage{algpseudocode}
\usepackage{wrapfig}

\newcommand{\cmark}{\ding{51}}
\newcommand{\xmark}{\ding{55}}

\definecolor{cvprblue}{rgb}{0.21,0.49,0.74}
\usepackage[pagebackref,breaklinks,colorlinks,allcolors=cvprblue]{hyperref}

\title{ReFlowSET: Representation-Aligned Latent Flow Matching\\for SAR-to-EO Image Translation}

\author{Jeonghyeok Do\\
[0.3em]
KAIST\\
{\tt\small ehwjdgur0913@kaist.ac.kr}
\and
Seungchul Lee\\
[0.3em]
Stellarvision Inc.\\
{\tt\small leesc@stellarvision.kr}
\and
Munchurl Kim \footnotemark[1]\\
[0.3em]
KAIST\\
{\tt\small mkimee@kaist.ac.kr}
\and
\small{\url{https://kaist-viclab.github.io/ReFlowSET_site}}
}

\begin{document}
\maketitle

{
  \renewcommand{\thefootnote}%
    {\fnsymbol{footnote}}
  \footnotetext[1]{Corresponding author.}
}

\begin{abstract}
SAR-to-EO image translation aims to generate electro-optical (EO)
imagery from synthetic aperture radar (SAR) observations. Existing
latent diffusion approaches typically inherit a predetermined
autoencoder, although reconstruction fidelity can vary substantially
across codecs and modalities. Because the latent codec affects the round-trip preservation of both SAR conditions and EO targets, codec selection constitutes a fundamental design choice; nevertheless, existing methods largely rely on codecs pretrained on natural images. To remedy this, we introduce ReFlowSET, a conditional latent flow-matching framework that selects its codec through a joint SAR--EO reconstruction audit. Rather than inheriting a heavyweight pretrained generator, ReFlowSET trains a substantially smaller conditional DiT from scratch in the selected latent space, using dual-stream SAR conditioning followed by joint feature refinement. To provide semantic guidance for this from-scratch training, intermediate noisy-EO features are aligned with clean target-EO representations extracted by a frozen vision foundation model. This alignment is used only during training and introduces no additional inference cost. Experiments on QXS-SAROPT and SAR2Opt demonstrate state-of-the-art performance across diverse perceptual fidelity and distributional metrics. Code and pretrained weights are publicly available at \url{https://github.com/KAIST-VICLab/ReFlowSET}.
\end{abstract}    
\section{Introduction}

Synthetic aperture radar (SAR) provides reliable Earth observation regardless of illumination and under most weather conditions. However, its coherent imaging mechanism introduces speckle and geometric distortions, making SAR imagery less intuitive to interpret than electro-optical (EO) imagery. SAR-to-EO image translation (SET)~\cite{zhao2022comparative} aims to generate an interpretable EO representation from a corresponding SAR observation, supporting visual analysis when optical measurements are unavailable. This task remains challenging because SAR backscatter and EO reflectance describe fundamentally different physical properties, resulting in an inherently ambiguous cross-modal mapping.

\begin{figure*}[t]
    \centering
    \includegraphics[height=0.38\textwidth]{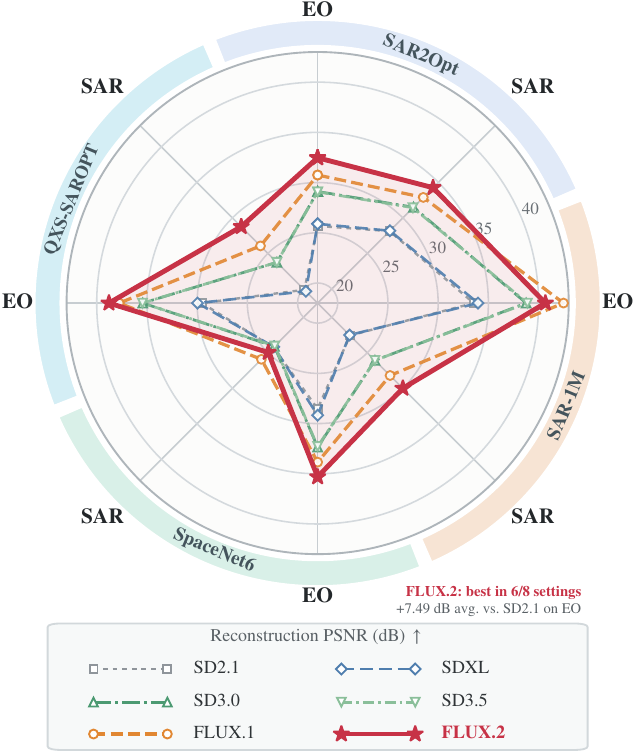}
    \hfill
    \includegraphics[height=0.38\textwidth]{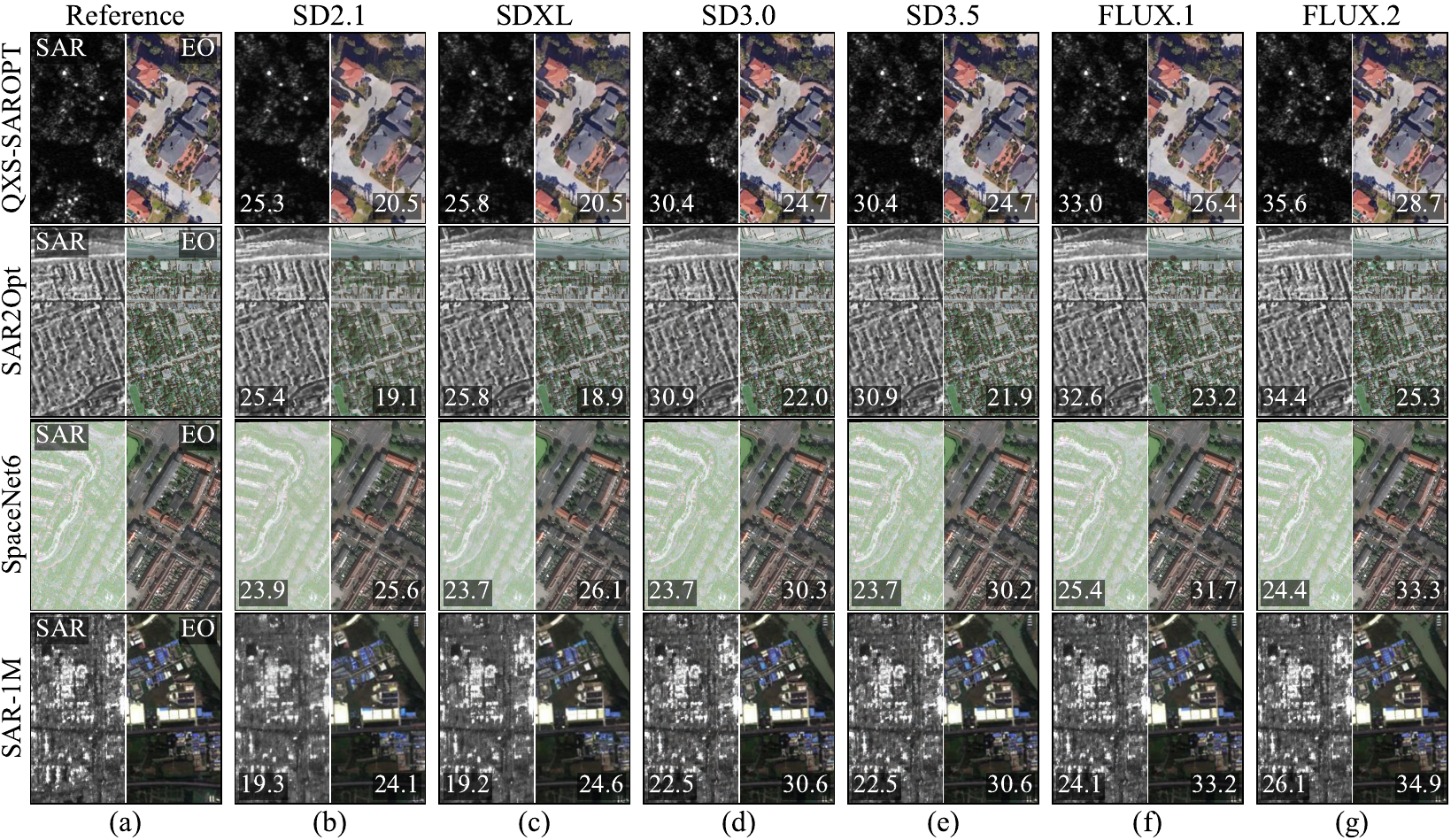}
    \caption{Modality-wise reconstruction ceilings of six pretrained autoencoders on four SAR--EO datasets. Left: encode--decode PSNR for SAR and EO imagery. Right: reference pairs and representative reconstructions, with SAR/EO PSNR values (dB) shown in each cell. Based on the dataset-level averages in the left panel, FLUX.2~\cite{flux-2-2025} provides the best reconstruction fidelity in six of eight dataset--modality settings and improves the mean EO reconstruction ceiling by 7.49 dB over SD2.1~\cite{rombach2022high}.}
    \label{fig:vae_ceiling}
\end{figure*}

Early SET approaches~\cite{turnes2020atrous, guo2024scene, lee2023segmentation, lee2023sar} primarily relied on generative adversarial networks (GANs), incorporating multiscale context, scene information, or semantic supervision. Although these methods enable direct translation, adversarial training may produce unstable textures and semantically incorrect objects. Diffusion-based approaches~\cite{bai2023conditional,qin2024efficient,kim2025conditional,do2026c} provide a more stable generative objective and improved perceptual quality. Recent latent methods~\cite{kim2025conditional,do2026c} further reduce computation but adopt the pretrained Stable Diffusion (SD) variational autoencoder (VAE)~\cite{rombach2022high} as a fixed codec without examining its suitability for SAR and EO imagery. Under latent-target training, the EO reconstruction represents a directly reachable decoded endpoint, while the SAR reconstruction indicates how much source structure is retained for conditioning. Moreover, existing latent SET models commonly inject SAR through immediate channel concatenation~\cite{do2026c}. This position-wise fusion may prematurely entangle modalities with different statistics, particularly when paired SAR--EO observations contain local spatial misalignment.

We introduce \textbf{ReFlowSET}, a \textbf{re}presentation-aligned latent \textbf{flow}-matching framework for \textbf{SET}. We first evaluate the modality-specific reconstruction ceilings of multiple pretrained VAEs~\cite{rombach2022high,podell2024sdxl,esser2024scaling,flux-2-2025} and select the codec providing the strongest joint fidelity for SAR and EO imagery. Rather than inheriting the selected codec's heavyweight pretrained generator, we train a substantially smaller conditional DiT from scratch using conditional flow matching~\cite{lipman2022flow}. The DiT employs dual-stream SAR conditioning that preserves modality-specific processing before channel-wise fusion and joint refinement. Because the from-scratch DiT does not inherit semantic priors from a pretrained generator, we align its intermediate noisy-EO features with clean-EO representations extracted by a frozen vision foundation model~\cite{simeoni2025dinov3}. This guidance is applied only during training and introduces neither additional annotations nor inference-time overhead.

The main contributions are summarized as follows:
\begin{itemize}
    \item We systematically evaluate multiple pretrained VAEs on SAR and EO imagery, revealing their modality-dependent reconstruction ceilings and identifying a suitable latent space for SET.
    \item We introduce a latent flow-matching framework with dual-stream SAR conditioning and training-only EO-derived VFM representation alignment.
    \item Extensive evaluations on QXS-SAROPT and SAR2Opt validate the proposed design, with ReFlowSET attaining leading performance in perceptual similarity and distributional realism.
\end{itemize}
\section{Related Work}

\subsection{General Image Translation and Restoration}
Conditional and cycle-consistent GANs established paired and unpaired image translation~\cite{isola2017image, zhu2017unpaired}; pix2pixHD~\cite{wang2018high} and SPADE~\cite{park2019semantic} improved high-resolution and spatially controlled synthesis. Diffusion methods later addressed super-resolution, generic translation, controlled generation, and restoration through SR3~\cite{saharia2022image}, BBDM~\cite{li2023bbdm}, ControlNet~\cite{zhang2023adding}, ResShift~\cite{yue2023resshift}, and HI-Diff~\cite{chen2023hierarchical}. StegoGAN~\cite{wu2024stegogan} targets non-bijective translation by separating hidden information. We retrain these families as broad image-to-image references rather than claiming that each is SET-specific.

\subsection{SAR-to-EO Translation}
SAR-specific GANs~\cite{turnes2020atrous,lee2023segmentation} incorporate atrous context or EO segmentation labels to reduce blur and semantic errors. Conditional Diffusion~\cite{bai2023conditional} performs iterative pixel-space denoising, whereas E3Diff~\cite{qin2024efficient} distills the process to one step. cBBDM~\cite{kim2025conditional} constructs a conditional Brownian bridge in the SD2.1 latent space, and C-DiffSET~\cite{do2026c} adds confidence-guided object generation to an SD2.1-based latent diffusion model. Our ReFlowSET instead treats the autoencoder endpoint as an explicit design variable and uses flow matching~\cite{lipman2022flow} with label-free, training-only representation guidance.
\section{Analysis of VAE Reconstruction Performance}
\label{sec:vae}

\begin{figure*}[tbp]
  \centering
  \includegraphics[width=1.0\textwidth]{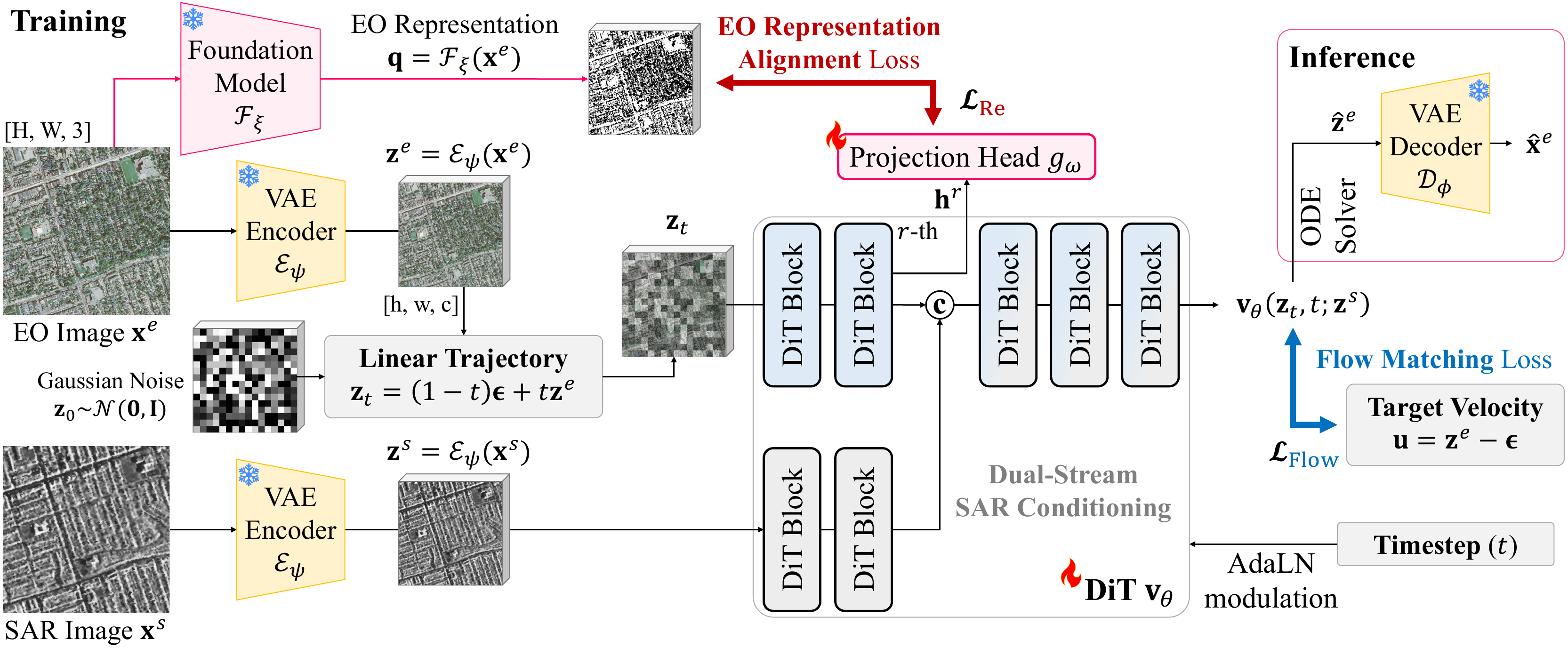}
  \caption{Overview of ReFlowSET. A conditional DiT learns the latent flow from Gaussian noise to the EO latent under dual-stream SAR conditioning. During training, intermediate DiT features are aligned with clean-EO representations from a frozen VFM. At inference, the alignment branch is removed and the integrated endpoint is decoded into an EO image.}
  \label{fig:reflowset}
\end{figure*}

Let $\mathcal{E}_{\psi}$ and $\mathcal{D}_{\phi}$ denote a frozen autoencoder. For an image $\mathbf{x}$, we define its \emph{round-trip reconstruction upper-bound} as the fidelity of $\mathcal{D}_{\phi}(\mathcal{E}_{\psi}(\mathbf{x}))$. Although this quantity is not a theoretical upper-bound over arbitrary latent codes, it represents a directly reachable endpoint under standard latent-target training: exact prediction of the encoded EO target produces this reconstruction. Accordingly, a higher EO upper-bound indicates less codec-induced target distortion and a more faithful attainable endpoint, while a higher SAR upper-bound implies that more source structure is retained for conditional generation. Autoencoder selection is therefore a consequential design choice rather than a fixed implementation detail.

Fig.~\ref{fig:vae_ceiling} compares the autoencoders of SD2.1~\cite{rombach2022high}, SDXL~\cite{podell2024sdxl}, SD3.0/3.5~\cite{esser2024scaling}, and FLUX.1/2~\cite{flux-2-2025} on QXS-SAROPT~\cite{huang2021qxs}, SAR2Opt~\cite{zhao2022comparative}, SpaceNet6~\cite{shermeyer2020spacenet}, and SAR-1M~\cite{liu2026sarmae} datasets. For each codec, dataset-level reconstruction statistics are computed only on the training splits. FLUX.2 achieves the highest PSNR in six of the eight dataset--modality settings and raises the mean EO ceiling by 7.49~dB over SD2.1, which is adopted by recent state-of-the-art latent SET methods~\cite{kim2025conditional, do2026c}. This reconstruction audit motivates the selection of FLUX.2 as a less distortion-limited latent codec. However, adopting its associated multi-billion-parameter pretrained generator would substantially increase computational cost. We therefore retain only the frozen FLUX.2 autoencoder for both SAR and EO in all subsequent experiments. To the best of our knowledge, this is the first systematic modality-wise analysis of pretrained autoencoder upper-bounds for SET.
\section{ReFlowSET}
\label{sec:method}

\begin{table*}[tbp]
    \scriptsize
    \centering
    \caption{
    Quantitative comparison on QXS-SAROPT~\cite{huang2021qxs}
    and SAR2Opt~\cite{zhao2022comparative}. All methods are retrained
    from official code on identical training splits and test pairs. \textbf{Bold} and \underline{underline} indicate the best and second-best results, respectively.}
    \label{tab:main}
    \resizebox{1.0\textwidth}{!}{%
    \def\arraystretch{1.2}
    \setlength{\tabcolsep}{5pt}
    \begin{tabular}{llcccccccccc}
    \toprule
    \multirow{2}{*}{\textbf{Method}}
    & \multirow{2}{*}{\textbf{Venue}}
    & \multicolumn{5}{c}{\textbf{QXS-SAROPT} dataset}
    & \multicolumn{5}{c}{\textbf{SAR2Opt} dataset} \\
    \cmidrule(lr){3-7} \cmidrule(lr){8-12}
    & & FID$\downarrow$ & DISTS$\downarrow$ & LPIPS$\downarrow$
      & SSIM$\uparrow$ & PSNR$\uparrow$
    & FID$\downarrow$ & DISTS$\downarrow$ & LPIPS$\downarrow$
      & SSIM$\uparrow$ & PSNR$\uparrow$ \\
    \midrule

    \multicolumn{12}{l}{\textit{General image translation/restoration methods:}}\\
    pix2pix~\cite{isola2017image}
      & CVPR'17
      & 174.6 & 0.373 & 0.665 & 0.203 & 12.33
      & 261.9 & 0.347 & 0.657 & 0.199 & 13.39 \\
    CycleGAN~\cite{zhu2017unpaired}
      & ICCV'17
      & 104.4 & 0.376 & 0.653 & 0.262 & 12.92
      & 143.5 & 0.330 & 0.650 & 0.178 & 12.90 \\
    pix2pixHD~\cite{wang2018high}
      & CVPR'18
      & 85.7 & 0.298 & 0.573 & 0.358 & 16.13
      & 146.3 & 0.283 & 0.567 & 0.268 & 15.95 \\
    SPADE~\cite{park2019semantic}
      & CVPR'19
      & 90.7 & 0.292 & 0.599 & 0.320 & 14.53
      & 142.5 & 0.265 & 0.597 & 0.234 & 14.47 \\
    DDPM (SR3)~\cite{saharia2022image}
      & TPAMI'22
      & 43.8 & 0.311 & 0.620 & 0.359 & 14.04
      & 122.5 & 0.295 & 0.610 & 0.313 & 13.65 \\
    SD2.1 fine-tune~\cite{rombach2022high}
      & CVPR'22
      & \textbf{19.1} & 0.257 & 0.561 & 0.348 & 15.40
      & \underline{71.8} & \underline{0.211} & 0.541 & 0.293 & 16.24 \\
    BBDM~\cite{li2023bbdm}
      & CVPR'23
      & 76.6 & 0.270 & 0.568 & 0.352 & 15.34
      & 143.1 & 0.290 & 0.590 & 0.276 & 15.29 \\
    ControlNet~\cite{zhang2023adding}
      & ICCV'23
      & 50.4 & 0.307 & 0.604 & 0.297 & 13.42
      & 140.5 & 0.350 & 0.643 & 0.217 & 11.73 \\
    ResShift~\cite{yue2023resshift}
      & NeurIPS'23
      & 140.2 & 0.334 & 0.607 & 0.217 & 14.20
      & 141.7 & 0.304 & 0.597 & 0.177 & 14.31 \\
    HI-Diff~\cite{chen2023hierarchical}
      & NeurIPS'23
      & 324.3 & 0.539 & 0.692 & \textbf{0.457} & \textbf{17.10}
      & 319.8 & 0.473 & 0.692 & \textbf{0.384} & \textbf{17.36} \\
    StegoGAN~\cite{wu2024stegogan}
      & CVPR'24
      & 106.8 & 0.384 & 0.658 & 0.254 & 12.96
      & 150.1 & 0.347 & 0.655 & 0.158 & 12.47 \\

    \midrule
    \multicolumn{12}{l}{\textit{SAR-to-EO methods:}}\\
    Cond. Diffusion~\cite{bai2023conditional}
      & GRSL'23
      & 88.6 & 0.355 & 0.730 & 0.213 & 11.55
      & 211.8 & 0.415 & 0.686 & 0.248 & 12.48 \\
    E3Diff~\cite{qin2024efficient}
      & GRSL'24
      & 47.8 & 0.278 & \underline{0.530} & 0.302 & 16.44
      & 104.7 & 0.232 & \underline{0.529} & 0.249 & 16.09 \\
    cBBDM~\cite{kim2025conditional}
      & GRSL'25
      & 50.6 & 0.246 & 0.539 & 0.372 & 16.02
      & 222.3 & 0.377 & 0.571
      & \underline{0.361} & \underline{17.05} \\
    C-DiffSET~\cite{do2026c}
      & TCSVT'26
      & \underline{19.9} & \underline{0.233} & \textbf{0.526}
      & \underline{0.380} & \underline{16.92}
      & 78.1 & 0.214 & \underline{0.529} & 0.314 & 16.81 \\
    \rowcolor{blue!5}
    ReFlowSET (SD2.1)
      & --
      & 25.5 & 0.267 & 0.566 & 0.337 & 15.85
      & 84.5 & 0.217 & 0.541
      & 0.270 & 15.82 \\
    \rowcolor{blue!5}
    \textbf{ReFlowSET (Ours)}
      & --
      & \textbf{19.1} & \textbf{0.231} & 0.534 & 0.355 & 16.09
      & \textbf{66.3} & \textbf{0.185} & \textbf{0.522}
      & 0.287 & 16.06 \\
    \bottomrule
    \end{tabular}}
\end{table*}

Fig.~\ref{fig:reflowset} illustrates the overall framework of our ReFlowSET, which combines conditional latent flow matching, dual-stream SAR conditioning, and training-only alignment with clean-EO representations.

\subsection{Conditional Latent Flow Matching}
For a paired SAR observation $\mathbf{x}^{s}$ and EO target $\mathbf{x}^{e}$, we encode the posterior means with the selected frozen encoder:
\begin{equation}
\mathbf{z}^{m}=\mathcal{E}_{\psi}(\mathbf{x}^{m}),
\quad m\in\{s,e\}.
\label{eq:encode}
\end{equation}
With $\boldsymbol{\epsilon}\!\sim\!\mathcal{N}(\mathbf{0},\mathbf{I})$ and $t\!\sim\!\mathcal{U}(0,1)$, a linear probability path $\mathbf{z}_t$ and its target velocity $\mathbf{u}$ are given by
\begin{equation}
\mathbf{z}_{t}=(1-t)\boldsymbol{\epsilon}+t\mathbf{z}^{e},
\qquad \mathbf{u}=\mathbf{z}^{e}-\boldsymbol{\epsilon}.
\label{eq:path}
\end{equation}
A conditional DiT~\cite{peebles2023scalable}, denoted as $\mathbf{v}_{\theta}$, receives $\mathbf{z}^{s}$ as a persistent condition, and is trained to minimize the flow matching loss:
\begin{equation}
\mathcal{L}_{\mathrm{Flow}}=
\mathbb{E}\!\left[\left\|
\mathbf{v}_{\theta}(\mathbf{z}_{t},t;\mathbf{z}^{s})-\mathbf{u}
\right\|_{2}^{2}\right].
\label{eq:fm}
\end{equation}
Starting from independent noise, rather than from the conditioned SAR latent $\mathbf{z}^s$, also prevents $\mathbf{u}$ from becoming an analytic function of the model inputs. At inference, the learned ordinary differential equation (ODE) is integrated from $t=0$ to $1$ and the endpoint is decoded by $\mathcal{D}_{\phi}$~\cite{lipman2022flow}.

\subsection{Dual-Stream SAR Conditioning}
A single-stream baseline~\cite{do2026c} concatenates
$\mathbf{z}^{s}$ and $\mathbf{z}_{t}$ channel-wise at the input, and processes them
through a shared DiT. In contrast, our ReFlowSET separately projects the
SAR condition and noisy-EO latent, and processes them using independently
parameterized streams for the first $r$ DiT blocks. The resulting
features are then concatenated along the channel dimension, projected
back to the model width, and are processed by the remaining single-stream
DiT blocks to predict the velocity field $\mathbf{v}_{\theta}(\mathbf{z}_{t},t;\mathbf{z}^{s})$.

\subsection{Training-Only EO Representation Alignment}
Because the DiT is trained from scratch, it does not inherit semantic priors from a pretrained generator. Moreover, flow matching supervises velocity prediction without directly constraining intermediate representations. We therefore use the clean EO target as a training-only representation teacher. A frozen VFM $\mathcal{F}_{\xi}$~\cite{simeoni2025dinov3} extracts representations $\mathbf{q}_{i}$, while a trainable projector $g_{\omega}$ maps the corresponding DiT features $\mathbf{h}_{i}^{r}$ into the same representation space as $\mathbf{q}_{i}$. Following the representation alignment~\cite{yu2024representation}, we minimize the cosine distance
\begin{equation}
\mathcal{L}_{\mathrm{Re}}
=
\frac{1}{N}\sum_{i=1}^{N}
\left(
1-
\frac{
g_{\omega}(\mathbf{h}_{i}^{r})^{\top}\mathbf{q}_{i}
}{
\|g_{\omega}(\mathbf{h}_{i}^{r})\|_{2}
\|\mathbf{q}_{i}\|_{2}
}
\right),
\end{equation}
where
$\mathbf{q}_{i}=\operatorname{sg}\!\left(\mathcal{F}_{\xi}(\mathbf{x}^{e})_{i}\right)$,
$\operatorname{sg}$ denotes stop-gradient,
$\mathbf{h}_{i}^{r}$ is the $i$-th spatial token of
the noisy-EO stream after the $r$-th dual-stream block, and $N$ is the number of aligned spatial tokens. The total training objective is
\begin{equation}
\mathcal{L}
=
\mathcal{L}_{\mathrm{Flow}} + \lambda\mathcal{L}_{\mathrm{Re}}.
\end{equation}
We empirically set $\lambda=0.5$ and keep it fixed across all experiments. The VFM $\mathcal{F}_{\xi}$ and projector $g_\omega$ are used only during training and discarded at inference, incurring no test-time overhead.

\begin{figure*}[tbp]
  \centering
  \includegraphics[width=1.0\textwidth]{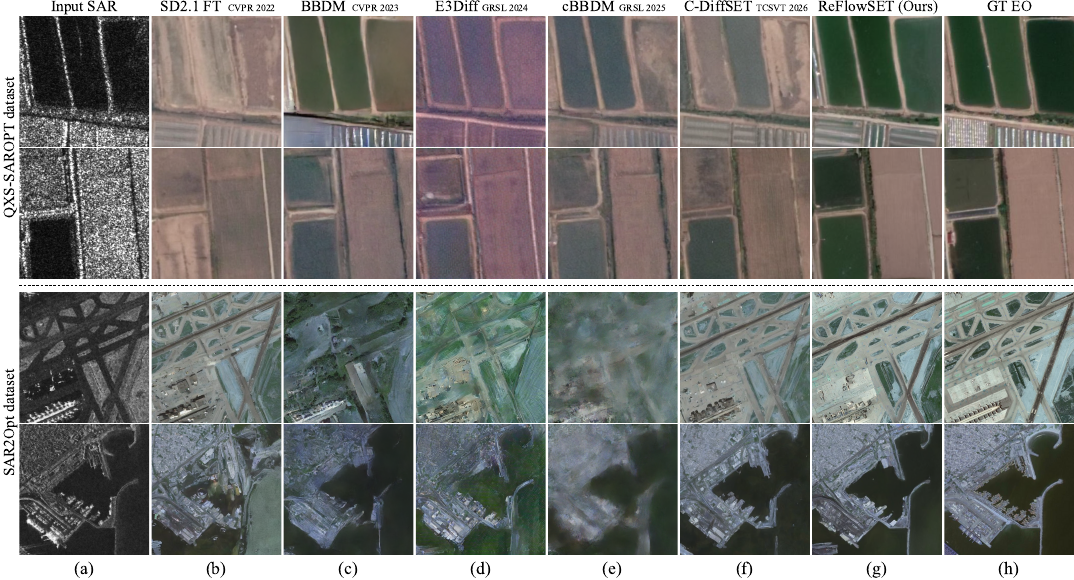}
  \caption{Qualitative SET comparison on QXS-SAROPT (top) and SAR2Opt (bottom). Columns show (a) input SAR, (b) SD2.1 FT, (c) BBDM, (d) E3Diff, (e) cBBDM, (f) C-DiffSET, (g) ReFlowSET, and (h) ground-truth EO.}
  \label{fig:main_result}
\end{figure*}

\begin{table*}[tbp]
    \scriptsize
    \centering
    \caption{Ablation of SAR conditioning, fusion, and EO representation
    alignment on SAR2Opt. All variants use 12 DiT blocks; D/S denote
    dual-/single-stream blocks. Params are deployed DiT parameters and
    exclude training-only alignment modules; VRAM and time report peak
    training memory and per-image inference latency.}
    \label{tab:dualstream_ablation}
    \setlength{\tabcolsep}{6pt}
    \renewcommand{\arraystretch}{1.2}
    \resizebox{0.70\textwidth}{!}{%
    \begin{tabular}{cccccccc}
        \toprule
        \makecell[c]{$\boldsymbol{\mathcal{L}}_{\mathrm{Re}}$}
        & \makecell[c]{\textbf{SAR}\\\textbf{Stream}}
        & \makecell[c]{\textbf{Fusion}}
        & \makecell[c]{\textbf{Topology}}
        & \makecell[c]{\textbf{FID}$\downarrow$}
        & \makecell[c]{\textbf{Params}$\downarrow$\\\textbf{(M)}}
        & \makecell[c]{\textbf{VRAM}$\downarrow$\\\textbf{(GB)}}
        & \makecell[c]{\textbf{Time}$\downarrow$\\\textbf{(sec)}} \\
        \midrule
        \cmark
        & Shared~\cite{do2026c}
        & Input~\cite{do2026c}
        & 0D+12S
        & 72.865
        & 108.46
        & 8.18
        & 0.92 \\
        
        \cmark
        & Dual
        & Token
        & 4D+8S
        & 70.450
        & 143.86
        & 13.57
        & 2.06 \\

        \xmark
        & Dual
        & Channel
        & 4D+8S
        & 71.851
        & 145.04
        & 9.80
        & 1.34 \\
        \rowcolor{blue!5}
        \cmark
        & Dual
        & Channel
        & 4D+8S
        & \textbf{70.436}
        & 145.04
        & 10.63
        & 1.34 \\
        \bottomrule
    \end{tabular}}
\end{table*}
\section{Experiments}

\subsection{Datasets and Settings}
QXS-SAROPT~\cite{huang2021qxs} contains 20,000 paired 256-pixel GF-3 SAR and Google Earth optical tiles; we use 16,001/3,999 images for training/testing. SAR2Opt~\cite{zhao2022comparative} contains 2,077 paired 600-pixel TerraSAR-X and Google Earth images; we use 1,450/627 images with random/center 512-pixel crops for training/testing. All baseline methods are retrained on the same training splits
and evaluated on identical test items. The randomly initialized DiT generator~\cite{peebles2023scalable} comprises 24 blocks and 509.3M parameters. The first $r=8$ blocks preserve separate modality-specific streams; EO representation alignment is applied to the EO-stream features at this boundary, after which the streams are concatenated and jointly processed by the remaining blocks. The FLUX.2~\cite{flux-2-2025} autoencoder and DINOv3 ViT-L/16~\cite{simeoni2025dinov3} teacher are pretrained and frozen. We train QXS-SAROPT for 40k updates with batch 64 and SAR2Opt for 20k updates with batch 32, respectively. All models were trained on two NVIDIA RTX 4090 GPUs. AdamW uses a peak learning rate of $5{\times}10^{-4}$, a 1k-step warmup, cosine decay, and bf16 precision. Conditioning dropout is 0.1. Reported samples use classifier-free guidance (CFG) of 1.5 and Euler solver with 50 function evaluations (NFE).

\subsection{Comparison with the State of the Art}
Table~\ref{tab:main} reports distributional realism (FID),
perceptual fidelity (DISTS and LPIPS), and pixel fidelity
(SSIM and PSNR). On QXS-SAROPT, ReFlowSET obtains the
best DISTS and ties the best FID. On SAR2Opt, it improves the
second-best FID from 71.8 to 66.3 and DISTS from 0.211 to
0.185, while also yielding the lowest (best) LPIPS. These reductions
correspond to relative gains of 7.7\% and 12.3\% in FID and
DISTS, respectively.

These results are particularly notable because SD2.1 fine-tuning
and C-DiffSET~\cite{do2026c} initialize their U-Net backbones
from pretrained SD2.1 weights, whereas ReFlowSET trains its
DiT generator from scratch. Within the same ReFlowSET framework,
replacing the SD2.1 codec with FLUX.2 improves all five metrics
on both benchmarks, reducing FID from 25.5 to 19.1 on
QXS-SAROPT and from 84.5 to 66.3 on SAR2Opt. This controlled
comparison empirically supports reconstruction-guided codec
selection: with FLUX.2, the from-scratch generator reaches or
surpasses pretrained-generator baselines on perceptual and
distributional metrics.

Although ReFlowSET does not maximize PSNR or SSIM, these
pixel-aligned metrics are sensitive to the inherent ambiguity
and local misalignment of paired SAR--EO observations. We
therefore emphasize perceptual and distributional metrics while
reporting pixel fidelity for completeness. Fig.~\ref{fig:main_result}
shows sharper structures and more coherent land-cover appearance
than competing latent diffusion and bridge models.

\subsection{Ablation and Efficiency Analysis}
Table~\ref{tab:dualstream_ablation} evaluates SAR conditioning, stream fusion, and EO representation alignment using compact 12-block models. To reduce ablation costs while controlling total depth, the dual-stream variants use 4 dual-stream and 8 single-stream blocks. The final ReFlowSET scales this allocation to 8 and 16 blocks, respectively, preserving the same 1:2 topology ratio.

\noindent \textbf{SAR conditioning and fusion.}
With EO representation alignment enabled, the dedicated SAR processing with channel fusion reduces FID from 72.865 to 70.436 relative to the shared-stream input-concatenation baseline~\cite{do2026c}. Because the modality-specific paths instantiate the same full-width DiT block design for both SAR and noisy EO, the dedicated conditioning increases the parameter count from 108.46M to 145.04M. This result should therefore be interpreted as a system-level architecture comparison. After the dual-stream stage, the token-wise fusion doubles the sequence length processed by every subsequent single-stream block. Consequently, it increases training VRAM from 10.63 to 13.57 GB and latency from 1.34 to 2.06 s/image, while providing virtually no FID benefit (70.450 versus 70.436). ReFlowSET therefore adopts channel-wise fusion.

\noindent \textbf{EO representation alignment.}
Under the same dedicated SAR stream, channel fusion, and 4D+8S topology, adding $\mathcal{L}_{\mathrm{Re}}$ reduces FID from 71.851 to 70.436. The deployed parameter count and inference latency remain unchanged because the frozen VFM and projection head are discarded after training. The alignment branch increases peak training VRAM only from 9.80 to 10.63 GB. This controlled comparison shows that target-EO representation alignment improves distributional quality without adding inference-time cost.
\section{Conclusion}

We presented ReFlowSET, a representation-aligned latent flow framework for SAR-to-EO translation. ReFlowSET combines reconstruction-guided codec selection, conditional latent flow matching, delayed SAR--EO fusion, and training-only EO representation alignment. It achieves the best FID and DISTS on both benchmarks and the best LPIPS on SAR2Opt among the evaluated methods, while the alignment branch introduces no inference-time overhead. Future work will examine cross-sensor generalization and the suppression of unsupported EO-like structures.

\vspace{0.2cm}
\noindent \textbf{Acknowledgement.}\quad This work was supported in part by the National Research Foundation of Korea (NRF) grant funded by the Korean government (MSIT) under the Sejong Science Fellowship Program (RS-2026-25484549, ``Generative AI-based High-Resolution SAR Image Visualization and Analysis Technology for All-Weather Earth Observation'', 50\%) and in part by the NRF grant funded by the MSIT (RS-2025-02222525, ``Development of AI-based SAR-to-EO image conversion technology'', 50\%).

\clearpage

{
    \small
    \bibliographystyle{ieeenat_fullname}
    \bibliography{main}

@String(PAMI  = {IEEE TPAMI})

@String(CVPR  = {CVPR})

@String(ICCV  = {ICCV})

@String(CVPRW = {CVPRW})

@String(ICLR  = {ICLR})

@String(ICML  = {ICML})

@String(NIPS  = {NeurIPS})

@String(GRSL  = {IEEE GRSL})

@String(TCSVT = {IEEE TCSVT})

@inproceedings{isola2017image,
  title     = {Image-to-image translation with conditional adversarial networks},
  author    = {Isola, Phillip and Zhu, Jun-Yan and Zhou, Tinghui and Efros, Alexei A.},
  booktitle = CVPR,
  pages     = {5967--5976},
  year      = {2017}
}

@inproceedings{zhu2017unpaired,
  title     = {Unpaired image-to-image translation using cycle-consistent adversarial networks},
  author    = {Zhu, Jun-Yan and Park, Taesung and Isola, Phillip and Efros, Alexei A.},
  booktitle = ICCV,
  pages     = {2242--2251},
  year      = {2017}
}

@inproceedings{wang2018high,
  title     = {High-resolution image synthesis and semantic manipulation with conditional {GANs}},
  author    = {Wang, Ting-Chun and Liu, Ming-Yu and Zhu, Jun-Yan and Tao, Andrew and Kautz, Jan and Catanzaro, Bryan},
  booktitle = CVPR,
  pages     = {8798--8807},
  year      = {2018}
}

@inproceedings{park2019semantic,
  title     = {Semantic image synthesis with spatially-adaptive normalization},
  author    = {Park, Taesung and Liu, Ming-Yu and Wang, Ting-Chun and Zhu, Jun-Yan},
  booktitle = CVPR,
  pages     = {2332--2341},
  year      = {2019}
}

@article{saharia2022image,
  title   = {Image super-resolution via iterative refinement},
  author  = {Saharia, Chitwan and Ho, Jonathan and Chan, William and Salimans, Tim and Fleet, David J. and Norouzi, Mohammad},
  journal = PAMI,
  volume  = {45},
  number  = {4},
  pages   = {4713--4726},
  year    = {2022}
}

@inproceedings{li2023bbdm,
  title     = {{BBDM}: Image-to-image translation with {Brownian} bridge diffusion models},
  author    = {Li, Bo and Xue, Kaitao and Liu, Bin and Lai, Yu-Kun},
  booktitle = CVPR,
  pages     = {1952--1961},
  year      = {2023}
}

@inproceedings{zhang2023adding,
  title     = {Adding conditional control to text-to-image diffusion models},
  author    = {Zhang, Lvmin and Rao, Anyi and Agrawala, Maneesh},
  booktitle = ICCV,
  pages     = {3813--3824},
  year      = {2023}
}

@article{yue2023resshift,
  title   = {{ResShift}: Efficient diffusion model for image super-resolution by residual shifting},
  author  = {Yue, Zongsheng and Wang, Jianyi and Loy, Chen Change},
  journal = NIPS,
  volume  = {36},
  pages   = {13294--13307},
  year    = {2023}
}

@article{chen2023hierarchical,
  title   = {Hierarchical integration diffusion model for realistic image deblurring},
  author  = {Chen, Zheng and Zhang, Yulun and Liu, Ding and Gu, Jinjin and Kong, Linghe and Yuan, Xin and others},
  journal = NIPS,
  volume  = {36},
  pages   = {29114--29125},
  year    = {2023}
}

@inproceedings{wu2024stegogan,
  title     = {{StegoGAN}: Leveraging steganography for non-bijective image-to-image translation},
  author    = {Wu, Sidi and Chen, Yizi and Mermet, Samuel and Hurni, Lorenz and Schindler, Konrad and Gonthier, Nicolas and Landrieu, Loic},
  booktitle = CVPR,
  pages     = {7922--7931},
  year      = {2024}
}

@article{turnes2020atrous,
  title   = {Atrous {cGAN} for {SAR}-to-optical image translation},
  author  = {Turnes, Javier Noa and Castro, Jose David Bermudez and Torres, Daliana Lobo and Vega, Pedro Juan Soto and Feitosa, Raul Queiroz and Happ, Patrick N.},
  journal = GRSL,
  volume  = {19},
  pages   = {1--5},
  year    = {2020}
}

@article{zhao2022comparative,
  title   = {A comparative analysis of {GAN}-based methods for {SAR}-to-optical image translation},
  author  = {Zhao, Yitao and Celik, Turgay and Liu, Nanqing and Li, Heng-Chao},
  journal = GRSL,
  volume  = {19},
  pages   = {1--5},
  year    = {2022}
}

@article{lee2023sar,
  title   = {{SAR}-to-virtual optical image translation for improving {SAR} automatic target recognition},
  author  = {Lee, In Ho and Park, Chan Gook},
  journal = GRSL,
  volume  = {20},
  pages   = {1--5},
  year    = {2023}
}

@article{guo2024scene,
  title   = {Scene-embedded generative adversarial networks for semi-supervised {SAR}-to-optical image translation},
  author  = {Guo, Zhe and Luo, Rui and Cai, Qinglin and Liu, Jiayi and Zhang, Zhibo and Mei, Shaohui},
  journal = GRSL,
  volume  = {21},
  pages   = {1--5},
  year    = {2024}
}

@article{lee2023segmentation,
  title   = {Segmentation-guided context learning using {EO} object labels for stable {SAR}-to-{EO} translation},
  author  = {Lee, Jaehyup and Kim, Hyun-Ho and Seo, Doochun and Kim, Munchurl},
  journal = GRSL,
  volume  = {21},
  pages   = {1--5},
  year    = {2023}
}

@article{bai2023conditional,
  title   = {Conditional diffusion for {SAR}-to-optical image translation},
  author  = {Bai, Xinyu and Pu, Xinyang and Xu, Feng},
  journal = GRSL,
  volume  = {21},
  pages   = {1--5},
  year    = {2023}
}

@article{kim2025conditional,
  title   = {Conditional {Brownian} bridge diffusion model for {VHR} {SAR}-to-optical image translation},
  author  = {Kim, Seon-Hoon and Chung, Daewon},
  journal = GRSL,
  volume  = {22},
  pages   = {1--5},
  year    = {2025}
}

@article{qin2024efficient,
  title   = {Efficient end-to-end diffusion model for one-step {SAR}-to-optical translation},
  author  = {Qin, Jiang and Zou, Bin and Li, Haolin and Zhang, Lamei},
  journal = GRSL,
  volume  = {22},
  pages   = {1--5},
  year    = {2024}
}

@article{do2026c,
  title   = {{C-DiffSET}: Leveraging latent diffusion for {SAR}-to-{EO} image translation with confidence-guided reliable object generation},
  author  = {Do, Jeonghyeok and Lee, Jaehyup and Lee, Seungchul and Kim, Munchurl},
  journal = TCSVT,
  year    = {2026}
}

@inproceedings{rombach2022high,
  title     = {High-resolution image synthesis with latent diffusion models},
  author    = {Rombach, Robin and Blattmann, Andreas and Lorenz, Dominik and Esser, Patrick and Ommer, Bj{\"o}rn},
  booktitle = CVPR,
  pages     = {10674--10685},
  year      = {2022}
}

@inproceedings{podell2024sdxl,
  title     = {{SDXL}: Improving latent diffusion models for high-resolution image synthesis},
  author    = {Podell, Dustin and English, Zion and Lacey, Kyle and Blattmann, Andreas and Dockhorn, Tim and M{\"u}ller, Jonas and Penna, Joe and Rombach, Robin},
  booktitle = ICLR,
  year      = {2024}
}

@inproceedings{esser2024scaling,
  title     = {Scaling rectified flow transformers for high-resolution image synthesis},
  author    = {Esser, Patrick and Kulal, Sumith and Blattmann, Andreas and Entezari, Rahim and M{\"u}ller, Jonas and Saini, Harry and Levi, Yam and Lorenz, Dominik and Sauer, Axel and Boesel, Frederic and Podell, Dustin and Dockhorn, Tim and English, Zion and Rombach, Robin},
  booktitle = ICML,
  series    = {Proceedings of Machine Learning Research},
  volume    = {235},
  pages     = {12606--12633},
  year      = {2024}
}

@misc{flux-2-2025,
  author       = {{Black Forest Labs}},
  title        = {{FLUX.2: Frontier Visual Intelligence}},
  howpublished = {\url{https://bfl.ai/blog/flux-2}},
  year         = {2025}
}

@inproceedings{lipman2022flow,
  title     = {Flow matching for generative modeling},
  author    = {Lipman, Yaron and Chen, Ricky T. Q. and Ben-Hamu, Heli and Nickel, Maximilian and Le, Matt},
  booktitle = ICLR,
  year      = {2023}
}

@inproceedings{peebles2023scalable,
  title     = {Scalable diffusion models with transformers},
  author    = {Peebles, William and Xie, Saining},
  booktitle = ICCV,
  pages     = {4172--4182},
  year      = {2023}
}

@inproceedings{yu2024representation,
  title     = {Representation alignment for generation: Training diffusion transformers is easier than you think},
  author    = {Yu, Sihyun and Kwak, Sangkyung and Jang, Huiwon and Jeong, Jongheon and Huang, Jonathan and Shin, Jinwoo and Xie, Saining},
  booktitle = ICLR,
  year      = {2025}
}

@article{simeoni2025dinov3,
  title   = {{DINOv3}},
  author  = {Sim{\'e}oni, Oriane and Vo, Huy V. and Seitzer, Maximilian and Baldassarre, Federico and Oquab, Maxime and Jose, Cijo and Khalidov, Vasil and Szafraniec, Marc and Yi, Seungeun and Ramamonjisoa, Micha{\"e}l and others},
  journal = {arXiv preprint arXiv:2508.10104},
  year    = {2025}
}

@article{huang2021qxs,
  title   = {The {QXS-SAROPT} dataset for deep learning in {SAR}-optical data fusion},
  author  = {Huang, Meiyu and Xu, Yao and Qian, Lixin and Shi, Weili and Zhang, Yaqin and Bao, Wei and Wang, Nan and Liu, Xuejiao and Xiang, Xueshuang},
  journal = {arXiv preprint arXiv:2103.08259},
  year    = {2021}
}

@inproceedings{shermeyer2020spacenet,
  title     = {{SpaceNet 6}: Multi-sensor all-weather mapping dataset},
  author    = {Shermeyer, Jacob and Hogan, Daniel and Brown, Jason and Van Etten, Adam and Weir, Nicholas and Pacifici, Fabio and H{\"a}nsch, Ronny and Bastidas, Alexei and Soenen, Scott and Bacastow, Todd and others},
  booktitle = CVPRW,
  pages     = {768--777},
  year      = {2020}
}

@inproceedings{liu2026sarmae,
  title     = {{SARMAE}: Masked autoencoder for {SAR} representation learning},
  author    = {Liu, Danxu and Wang, Di and Wang, Hebaixu and Chen, Haoyang and Jiang, Wentao and Cheng, Yilin and Guo, Haonan and Cui, Wei and Zhang, Jing},
  booktitle = CVPR,
  pages     = {6496--6507},
  year      = {2026}
}
}


\end{document}